\documentclass[a4paper,fleqn,numbers]{cas-dc}

\usepackage[numbers]{natbib}
\usepackage{amsmath}
\usepackage{algorithmic}
\usepackage{algorithm}
\usepackage{subcaption}
\usepackage{graphicx}
\usepackage{caption}
\usepackage{url}
\usepackage{booktabs} 
\usepackage{multirow} 
\def\tsc#1{\csdef{#1}{\textsc{\lowercase{#1}}\xspace}}
\tsc{WGM}
\tsc{QE}
\begin{document}
\let\WriteBookmarks\relax
\def\floatpagepagefraction{1}
\def\textpagefraction{.001}
\let\printorcid\relax % 可去掉页面下方的ORCID(s)

% \shorttitle{<short title of the paper for running head>} 
 \shorttitle{Regression generation adversarial network based on dual data evaluation strategy for industrial soft sensing application}    

 \shortauthors{Wang}
% \shortauthors{V. {{\=A}}nand Rawat et al.}

\title[mode = title]{Regression generation adversarial network based on dual data evaluation strategy for industrial soft sensing application}

\author{Zesen Wang$^a$} \author{Yonggang Li$^a$}\author{Lijuan Lan$^{a,b,*}$}
\affiliation[aff1]{organization={School of Automation},
	addressline={Central South University}, 
	city={Changsha},
	postcode={410083}, 
	state={Hunan},
	country={China}}
\affiliation[aff2]{organization={State Key Laboratory of Precision Manufacturing for Extreme Service Performance},
	addressline={Central South University}, 
	city={Changsha},
	postcode={410083}, 
	state={Hunan},
	country={China}}

\cortext[cor1]{Corresponding author}

\begin{abstract}
Soft sensing infers hard-to-measure data through a large number of easily obtainable variables. However, in complex industrial scenarios, the issue of insufficient data volume persists, which diminishes the reliability of soft sensing. Generative Adversarial Networks (GAN) are one of the effective solutions for addressing insufficient samples. Nevertheless, traditional GAN fail to account for the mapping relationship between labels and features, which limits further performance improvement. Although some studies have proposed solutions, none have considered both performance and efficiency simultaneously. To address these problems, this paper proposes the multi-task learning-based regression GAN framework that integrates regression information into both the discriminator and generator, and implements a shallow sharing mechanism between the discriminator and regressor. This approach significantly enhances the quality of generated samples while improving the algorithm's operational efficiency. Moreover, considering the importance of training samples and generated samples, a dual data evaluation strategy is designed to make GAN generate more diverse samples, thereby increasing the generalization of subsequent modeling. The superiority of method is validated through four classic industrial soft sensing cases: wastewater treatment plants, surface water, $CO_2$ absorption towers, and industrial gas turbines.
\end{abstract}


%\item highlight-1
%\item highlight-2
%\item highlight-3

% Keywords

\maketitle

% Main text
\section{Introduction}
\par With the acceleration of industrialization, environmental pollution has become increasingly severe. For instance, wastewater discharged from some sewage treatment facilities does not meet national regulatory standards, which not only damages the ecological environment but also poses potential threats to public health \cite{li2024multi}. Greenhouse gases such as carbon dioxide emitted by process industries are among the significant factors contributing to global warming \cite{gong2024dynamic}. The existence of these problems seriously hinders the development of modern industrial parks towards intelligence and sustainability. Accurate perception of environmental pollution is a prerequisite for implementing prevention and control measures; however, traditional methods suffer from problems such as long processing times and low accuracy. Against this backdrop, data-driven soft sensing methods gradually emerge \cite{geng2023self}. Researchers predict hard-to-measure quality variables through easily measurable process variables. A substantial body of related research demonstrates the effectiveness of soft sensing methods \cite{10912639}.
\par Relevant studies indicate that data-driven soft sensing methods require a large number of labeled samples to ensure the accuracy of predictions. However, for the majority of industrial processes, it is not feasible to guarantee a sufficient volume of data. Currently, some scholars attempt to address this problem through Semi-Supervised Learning (SSL) and Generative Adversarial Networks (GAN) \cite{song2022graph,gui2021review,qin2024improved}. SSL is a method that leverages a large quantity of unlabeled samples to enhance model performance when labeled samples are scarce \cite{yang2022survey}. Unfortunately, in many industrial settings, obtaining even a substantial amount of unlabeled samples become a luxury, making it difficult to collect enough data to establish SSL model. GAN circumvent this drawback by being a technology that directly generates additional pseudo-samples \cite{zhou2022distribution}. The architecture of GAN comprises a discriminator and a generator, which engage in continuous competition during the training process \cite{yang2024novel}. Ultimately, the generator becomes capable of producing pseudo-samples that are nearly indistinguishable from real samples. Currently, many scholars apply GAN to the modeling of soft sensors \cite{wang20252d}. For example, Zhu et al. \cite{zhu2022partial} use GAN for data augmentation of partial discharge data, effectively mitigating overfitting caused by low data volume or imbalanced distributions; Zhao et al. \cite{zhao2022state} apply GAN to estimate the state of health of batteries, ensuring that the effectiveness of soft sensing models is no longer limited by sample size.
\par Currently, in the soft sensing applications of GAN, the common practice is to concatenate labels with features as a whole input for training the GAN, where the newly generated labels are merely one dimension of the generated data \cite{deng2025time}. However, this approach overlooks the regression relationship between labels and features, hindering further improvements in the performance of regression models using the generated data. It may even generate erroneous data, leading to a decrease in model accuracy. A few studies improve upon this by incorporating regressor into the data generation process to integrate the regression relationship between labels and features \cite{pham2022conditional}. For example, Li et al. \cite{li2023data} propose MR-GAN, which introduces regression information into the generator and integrates modal information as a condition within both the generator and discriminator to guide the generation of multi-modal regression data. Meanwhile, Jiang et al. \cite{jiang2022ragan} propose RA-GAN, introducing a regression attention mechanism into the parameter learning of both the discriminator and generator. 
\par The aforementioned methods have all taken into account the regression relationship between labels and features, achieving a certain level of success. However, there are still some limitations. MR-GAN only integrates regression information within the generator without considering the discriminator. Moreover, it re-trains the regressor using both real and generated samples in each adversarial learning iteration, which can enhance the quality of the generated data but requires a substantial amount of training time. RA-GAN embeds regression relationships in both the generator and discriminator but fails to adequately consider the similarity between the tasks of the discriminator and regressor, which hinders further performance and  efficiency improvements.
\par Given the drawbacks of the aforementioned methods, considering that iterative optimization of the regressor can consume a large amount of computational resources, while neglecting this optimization can hinder the quality of the generated data, we propose a  multi-task learning-based regression GAN framework (RGAN) that strikes a balance between these two extremes. Firstly, we incorporate the mapping relationship between labels and features into the generator's loss function to ensure that the generated samples carry regression information. Furthermore, recognizing the correlation between the tasks performed by the regressor and the discriminator, we implement a shallow sharing mechanism between them. A shared underlying network is responsible for capturing general features beneficial to all tasks, while top-level task-specific modules are adjusted according to their respective task requirements. This approach not only enhances the learning efficiency of the model but also contributes to improving its generalization ability. 
\par In addition to the design of GAN architecture, the quality assessment of training samples and generated samples is also crucial. Therefore, we propose a dual data evaluation strategy to assess and select the highest quality training and generated samples \cite{wang2021gan}. High-quality training samples enable GAN to generate more representative and diverse samples, while high-quality generated samples can effectively improve the accuracy and generalization of soft sensing models. For training samples, we employ active learning to select samples that are more uncertain, representative, and diverse for training the generator and discriminator. This approach can effectively reduce the cost of labeled data and guide the generator to produce higher-quality samples \cite{wu2019active,chen2022supervised}. For generated samples, we introduce a comprehensive evaluation framework to assess the quality of the generated samples, thereby selecting those that exhibit better consistency with the distribution of real samples and have a more diverse distribution. 
\par In summary, we propose a regression generative adversarial network based on dual data evaluation strategy (RGAN-DDE). The main contributions of this paper are as follows:
\begin{enumerate}
	\item We introduce a multi-task learning-based regression GAN framework, which allows for shallow sharing between the regressor and the discriminator. This ensures efficient model learning while enhancing the model's generalization capability. Additionally, during the generator updates, regression information is backpropagated to ensure that the generated samples carry regression information.
	\item We propose a dual data evaluation strategy to filter out high-quality training and generated data. This reduces the cost of labeling data and improves the accuracy and generalization of subsequent modeling.
	\item Extensive experiments are conducted on four typical industrial datasets, thoroughly comparing our approach with state-of-the-art algorithms. The experimental results validate the effectiveness of the proposed method.
\end{enumerate}

\section{Related work}
\subsection{GAN, WGAN-GP, and MR-GAN}
\par
\begin{figure}
	\centering
	\includegraphics[width=0.9\linewidth]{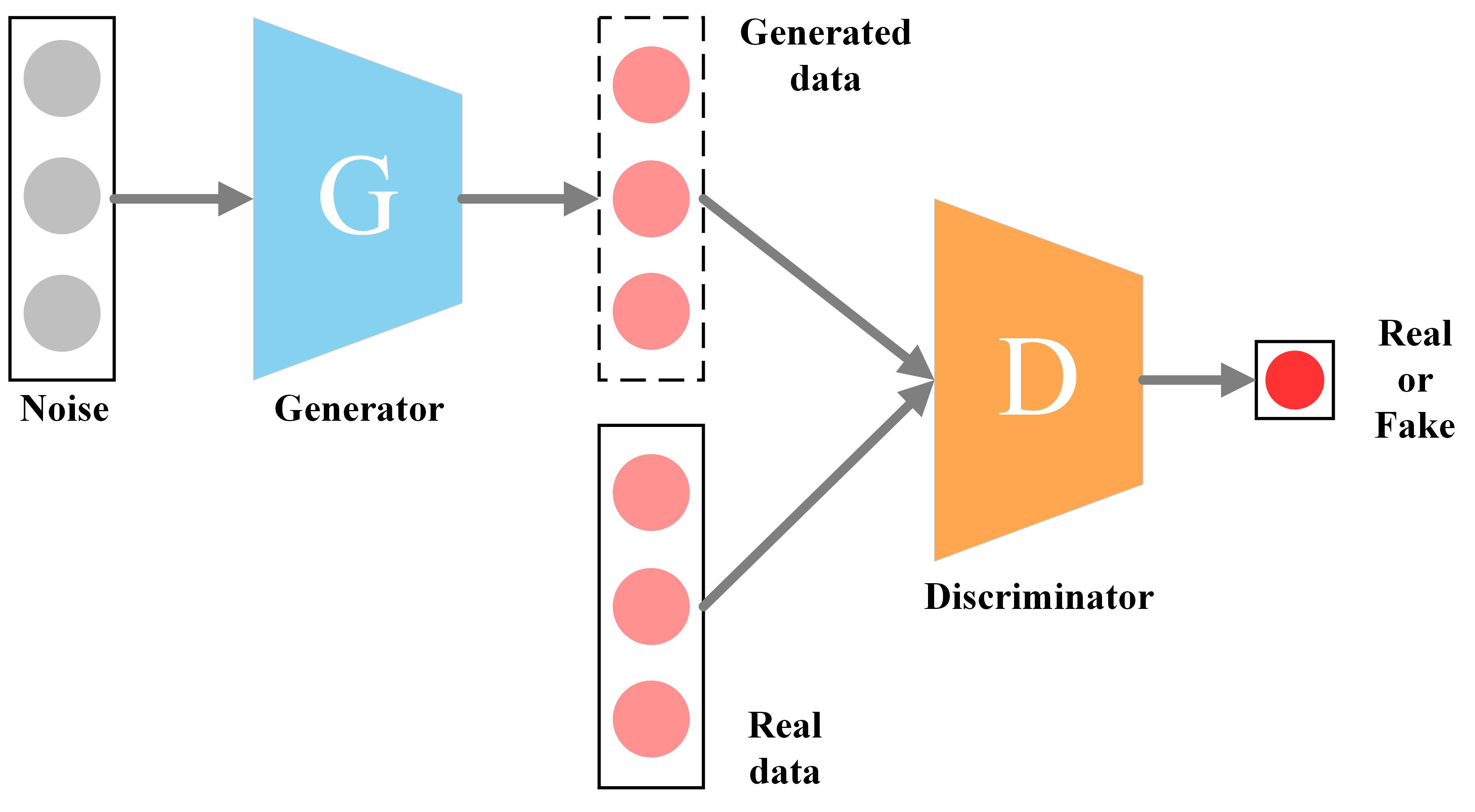}
	\caption{Structure of GAN.}
	\label{fig:GAN}
\end{figure}

 GAN is introduced by Goodfellow et al. \cite{goodfellow2014generative} in 2014, primarily comprising a generator G and a discriminator D, as illustrated in Figure \ref{fig:GAN}. The core idea behind GAN is to train the model through an adversarial process between G and D: G aims to generate samples that are as realistic as possible, while D strives to improve its capability to correctly classify samples, making it difficult to distinguish the distribution of generated data from that of real data. The objective function of GAN can be expressed as: 
\begin{equation}
	\begin{aligned}
		\min_G \max_D V(D,G) &= \mathbb{E}_{x \sim p_r}[\log D(x)] 
		&\quad \\&+ \mathbb{E}_{z \sim p_z}[\log (1 - D(G(z)))],
		\label{eq1}
	\end{aligned}
\end{equation}
where \( z \) is the random noise, \( p_r \) is the real data distribution, and \( p_z \) is the noise distribution input to the generator.
\par However, GAN suffer from problems of training instability and mode collapse \cite{arjovsky2017wasserstein}. To address these problems, Wasserstein GAN (WGAN) introduces the Wasserstein distance to replace the Jensen-Shannon divergence for measuring the similarity between two distributions. This provides a meaningful gradient for updating the network parameters, with the objective function formulated as follows: 
\begin{equation}
	W(p_r, p_g) = \sup_{||f||_L \le 1} \mathbb{E}_{x \sim p_r}[f(x)] - \mathbb{E}_{x' \sim p_g}[f(x')],
	\label{eq2}
\end{equation}
where \(p_r\) represents the real data distribution, \(p_g\) represents the generated data distribution, and \(f\) is a discriminator chosen from a specific class of function spaces. 
\par To ensure that \(f\) satisfies the Lipschitz condition, WGAN clips the discriminator's parameters, which can lead to optimization difficulties. To address the training challenges of WGAN, Arjovsky proposes Wasserstein GAN with Gradient Penalty (WGAN-GP) \cite{gulrajani2017improved}. In WGAN-GP, gradient penalty is used instead of weight clipping, further enhancing the stability of model training. The objective function for WGAN-GP is expressed as: 
% Requires: \usepackage{amsmath}
\begin{equation}
	\begin{aligned}
		\min_{G} \max_{D} V(D, G) &= \mathbb{E}_{x \sim p_r}[D(x)] - \mathbb{E}_{x \sim p_g}[D(x)] \\&+ \lambda \mathbb{E}_{x \sim p}[(\| \nabla_x D(x) \|_2 - 1)^2],
		\label{eq3}
	\end{aligned}
\end{equation}
where \(\lambda\) is the penalty coefficient, and $x\sim p$ represents that the data distribution comes from \(p_r\) and \(p_g\). 
\par However, some studies indicate that directly applying WGAN-GP to soft sensing applications does not yield good results, because WGAN-GP does not consider the regression mapping relationship between features and labels; it merely treats labels as an additional dimension of the data. To address this problem, Li et al. \cite{li2023data} propose MR-GAN, which incorporates regression information into the parameter updates of the generator. The objective function for the generator is presented as follows: 
\begin{equation}
	L_G = - \frac{1}{N}\sum\limits_{i = 1}^N \left( D(G(z)) + \frac{\alpha }{N} \sum\limits_{i = 1}^N \left( (\hat{y}_i - y_i)^2 \right) \right),
	\label{eq4}
\end{equation}
where \(\alpha\) is the weight for sparsity, \(\hat{y}_i\) is the predicted value, and \(y_i\) is the true value. 
\par Although MR-GAN further improves the performance of WGAN-GP in soft sensing applications, it does not incorporate regression information into the discriminator, which limits the potential for further performance improvement. Moreover, during each iteration of optimization, the regressor is retrained using both real and generated samples. While this can improve the quality of the generated data, it requires a significant amount of training time. Therefore, inspired by MR-GAN and RA-GAN \cite{jiang2022ragan, li2023data}, we incorporate regression information into both the generator and the discriminator. Exploiting the architectural similarity between the regressor and discriminator, we implement an information sharing mechanism to enhance training efficiency \cite{zhao2022state}. Beyond data quantity, we further emphasize data quality by proposing a dual data evaluation strategy.

\begin{figure}
	\centering
	\includegraphics[width=0.8\linewidth]{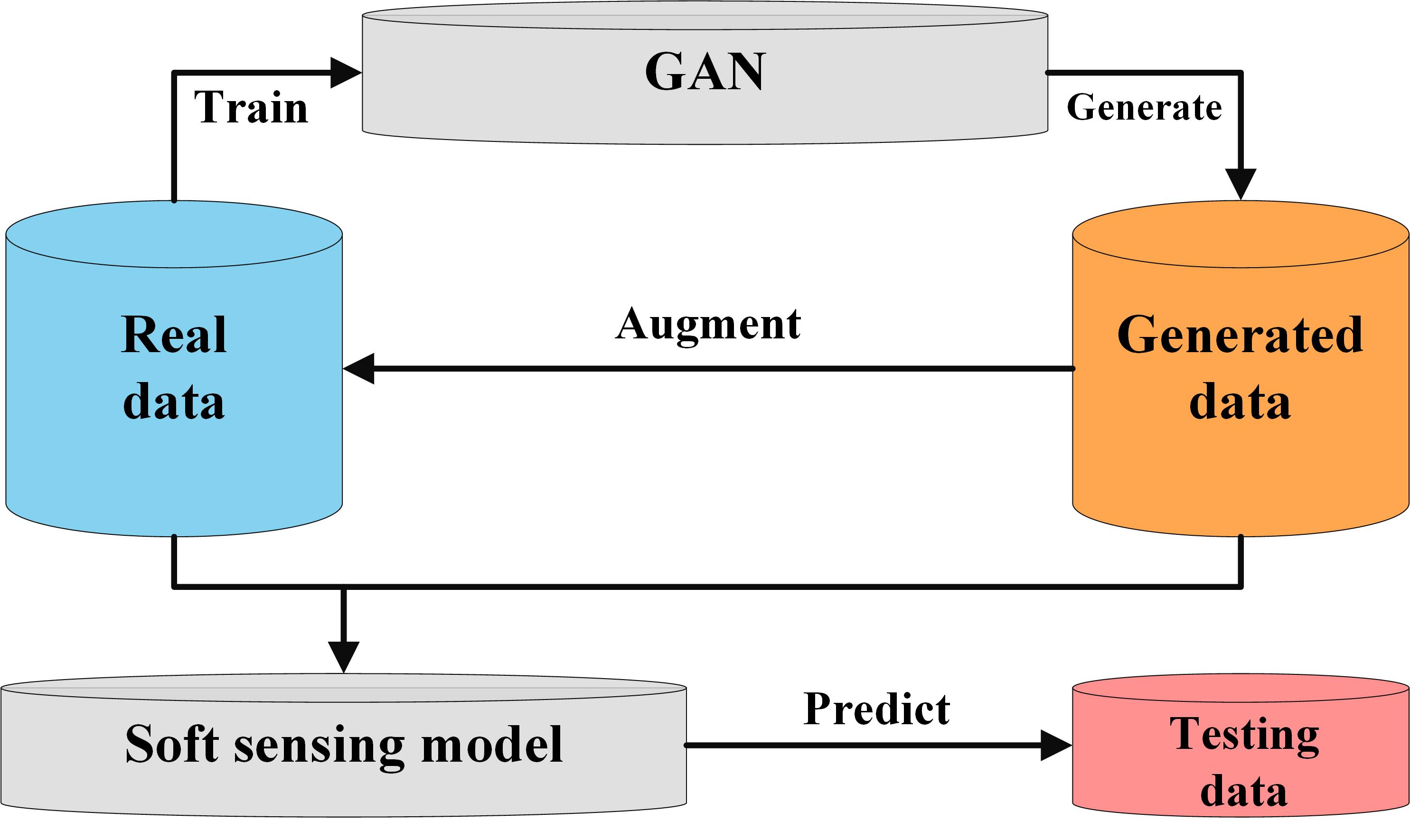}
	\caption{Flowchart of GAN for soft sensor.}
	\label{fig:soft sensor}
\end{figure}

\subsection{GAN for Soft Sensor}
\par In industrial soft sensing applications \cite{11053977, rezagholiradeh2018reg}, a data augmentation framework based on GAN-like models is illustrated in Figure \ref{fig:soft sensor}. Since GAN is an unsupervised model, we need to construct the input samples prior to training. Assuming the features of real data are represented by ${x_i} = [x_i^{(1)},x_i^{(2)},...,x_i^{(d)}] \in X = \{ {x_1},{x_2},...,{x_M}\} $, where $d$ is the data dimension, and the labels of the real data are denoted by ${y_i} \in Y = \{ {y_1},{y_2},...,{y_M}\}$. Then, the input to the GAN is constructed as $[{x_i},{y_i}] \in D = \{X,Y\}$. The main steps are outlined as follows:

\begin{enumerate}
	\item Train the GAN using dataset $D$.
	\item Generate new data $[{x'_i},{y'_i}] \in D' = \{X',Y'\}$ using the trained GAN.
	\item Train a regression model using $\{ X,X'\} $ as features and $\{ Y,Y'\} $ as labels.
	\item Input the test data into the trained regression model to obtain the predicted values.
\end{enumerate}

\section{Propoesd method}
\begin{figure*}
	\centering
	\includegraphics[width=1\linewidth]{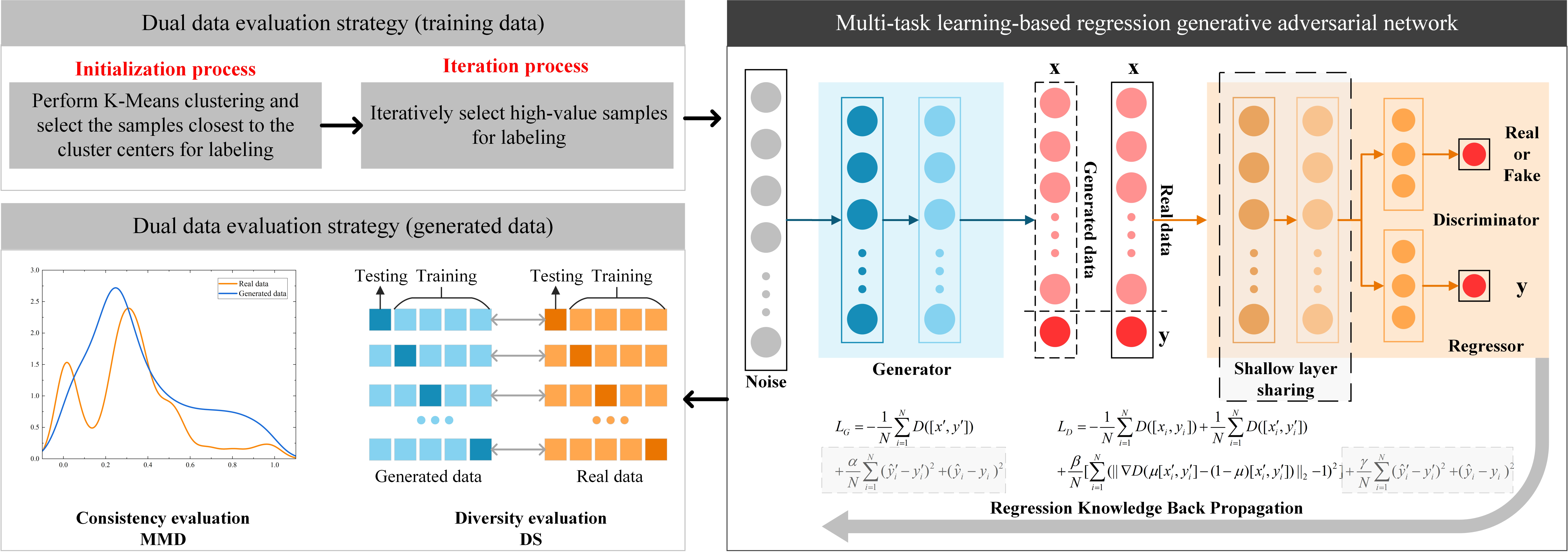}
	\caption{Structure of RGAN-DDE}
	\label{fig:RGAN-DDE}
\end{figure*}

In practical industrial production environments, ensuring sufficient industrial data volume often poses a challenge. Therefore, utilizing data augmentation techniques to expand the dataset becomes crucial. Considering the drawbacks of existing methods, we propose a regression generative adversarial neural network based on a dual data evaluation strategy (RGAN-DDE), which primarily consists of two parts: multi-task learning-based regression GAN (RGAN) and dual data evaluation strategy, as shown in Figure \ref{fig:RGAN-DDE}. For RGAN, we incorporate regression information into both the generator and discriminator and implement a shallow sharing mechanism between the regressor and discriminator. This approach ensures model performance while significantly enhancing training efficiency. For the dual data evaluation strategy, there are two main components: acquiring high-quality training samples to enable RGAN to generate more diverse samples, and evaluating the quality of the generated samples to ensure the accuracy and generalization of subsequent soft measurement modeling. The following discusses this in detail.
\subsection{RGAN}
\par To incorporate the causal relationship between labels and features within the WGAN-GP framework, we integrate a regression model (regressor) into this framework. The regressor employs a deep network model and is trained with pre-training to ensure stable training. Both generated data and real data are fed into the regressor to improve its generalization ability. The objective function is as follows: 
\begin{equation}
	L_R = \frac{1}{N}\sum\limits_{i = 1}^N \left( (\hat{y'}_i - y'_i)^2 + (\hat{y}_i - y_i)^2 \right),
	\label{eq5}
\end{equation}
where \(\hat{y}_i\) represents the predicted value for real samples, \(y_i\) represents the observed value for real samples, \(\hat{y'}_i\) represents the predicted value for generated samples, and \({y'}_i\) represents the observed value for generated samples. 
\par Furthermore, the regression information is incorporated into the back propagation process of the generator to ensure that there is an appropriate mapping relationship between the labels and features generated by the generator. The objective function for the generator is presented as follows: 
\begin{equation}
	\begin{aligned}
		{L_G} =  - \frac{1}{N}\sum\limits_{i = 1}^N {D([x',y']) + \frac{\alpha }{N}} \sum\limits_{i = 1}^N {{{({{\hat y'}_i} - {{y'}_i})}^2} + } {({\hat y_i} - y_i^{})^2},
		\label{eq6}
	\end{aligned}
\end{equation}
where \((x',y')\) is the generated sample, \(\alpha\) is a hyperparameter.
\par To avoid the drawbacks of existing methods and considering the similarity in tasks performed by the regressor and discriminator, we implement a shallow sharing mechanism between the regressor and discriminator. Through this mechanism, not only is effective parameter sharing achieved, promoting collaborative learning between regression and classification tasks, but regression information is also seamlessly integrated into the discriminator. This approach not only improves training efficiency but also significantly enhances the quality of the generated samples. To this end, we define a joint objective function to simultaneously optimize both the discriminator and regressor: 
\begin{equation}
	\begin{aligned}
		L_D = & - \frac{1}{N}\sum\limits_{i=1}^N D([x_i, y_i]) + \frac{1}{N}\sum\limits_{i=1}^N D([x'_i, y'_i]) \\
		& + \frac{\beta}{N} \left[\sum\limits_{i=1}^N \left(||\nabla D(\mu [x'_i, y'_i] - (1 - \mu)[x'_i, y'_i])||_2 - 1\right)^2 \right] \\
		& + \frac{\gamma}{N} \sum\limits_{i=1}^N \left((\hat{y}'_i - y'_i)^2 + (\hat{y}_i - y_i)^2\right),
	\end{aligned}
	\label{eq7}
\end{equation}
where \(\beta, \gamma\) are hyperparameters.

\subsection{Dual Data Evaluation Strategy}
\par For GAN, the quality of both training samples and generated samples is crucial. For training samples, high-quality training data can achieve superior model performance even with smaller datasets. Moreover, it can guide the generator to produce diverse samples, avoid learning biased or limited patterns, and promote the generation of richer and more realistic samples, thereby significantly enhancing the generalization ability of subsequent regression modeling. For generated samples, their directly determines the prediction accuracy of follow-up modeling tasks. High-quality generated samples not only enhance the model's generalization capability but also provide better support for subsequent regression tasks. Therefore, we propose a dual data evaluation strategy to select high-quality training and generated samples. 
\par For the selection of training samples, we adopt an active learning strategy that comprehensively evaluates training samples based on diversity and representativeness to choose the most valuable samples for guiding the generator in producing data \cite{liu2020integrating}. Additionally, this strategy effectively reduces labeling costs, enabling the construction of high-performance models with fewer training samples. Assuming there are currently unlabeled training samples, the process begins with K-means clustering of all initial samples. The optimal number of clusters is determined using the average silhouette coefficient, and samples closest to the cluster centers are selected for manual annotation. Following this, the distances between the remaining samples and the cluster centers are calculated. Based on the cost of labeling, samples that are farther from the cluster centers are chosen for annotation. Through these steps, a more representative and diverse training dataset is constructed at a lower labeling cost. The pseudo code for this process is provided in the Algorithm \ref{active learning}.
\begin{algorithm}[h]
	\caption{Active learning algorithm}
	\label{active learning}
	\begin{algorithmic}[1]
		\STATE \textbf{Input:} A pool of $N$ unlabeled samples, $\{x_i\}_{i=1}^N$; $M_{max}$, the maximum number of samples to label.
		\STATE \textbf{Output:} The training dataset $D=\{X,Y\}$.
		
		\STATE \textbf{Initialization process:}
		\STATE Perform $k$-means ($k=M_0$) clustering on $\{x_i\}_{i=1}^N$.
		\FOR{each cluster}
		\STATE Select the samples closest to the cluster centers for labeling, and add them to the training dataset $D$.
		\ENDFOR
		\STATE Construct the initial regression model $f(x)$.
		\STATE \textbf{Iteration process:}
		\FOR{$M = M_0 + 1, \dots, M_{max}$}
		\STATE Compute $d^{xy}_{n}$ through $d_n^{xy} = d_n^{x} \cdot d_n^{y}$ for each unlabeled sample $\{x_n\}_{n=M}^N$, where $d_n^x = \min_{m} \| x_n - x_m \|, \quad m = 1, \ldots, M; \quad n = M + 1, \ldots, N$.
		\STATE Compute $R$ through $R_n = \sum_{i=1}^{N} || {x}_n - {x}_i ||$ for each unlabeled sample $\{x_n\}_{n=M}^N$.
		\STATE Use $x = \arg\max\limits_{x_n} ( \frac{d_n^{xy}}{R_n}), \quad n = M+1, \ldots, N$ to select one sample for labeling, and add it to the training dataset $D=\{X,Y\}$.
		\ENDFOR
	\end{algorithmic}
\end{algorithm}

\par For the evaluation of generated samples, a good sample augmentation model should enhance sample diversity while effectively simulating the distribution of actual samples, thereby improving the predictive performance and generalization ability of subsequent modeling. To this end, we assess the diversity and consistency of the generated samples. In this paper, we use Maximum Mean Discrepancy (MMD) to estimate the distributional differences between generated samples and real samples. A lower MMD value indicates higher similarity between the two distributions. The mathematical formulation is described as follows: 
\begin{equation}
	\begin{aligned}
		L_{MMD}^2 = & \frac{1}{n^2}\sum_{i = 1}^n \sum_{i' = 1}^n k([x_i, y_i],[x_{i'}, y_{i'}]) \\
		& - \frac{2}{nm}\sum_{i = 1}^n \sum_{j = 1}^m k([x_i, y_i],[x'_j, y'_j]) \\
		& + \frac{1}{m^2}\sum_{j = 1}^m \sum_{j' = 1}^m k([x_j, y_j],[x'_{j'}, y'_{j'}]),
	\end{aligned}
	\label{eq8}
\end{equation}
where \(k\) is the kernel mapping function, and \(n, m\) are the number of real samples and generated samples, respectively.
\par On the other hand, the diversity of generated samples is also important as it contributes to improving the generalization performance of subsequent modeling. Therefore, this paper introduces a Diversity Score (DS) to estimate the diversity of generated samples. We employ cross-validation between the generated dataset and the real dataset to assess the generalization performance. Specifically, both the generated and real datasets are each divided into $K$ equal parts. Then, a regression model is trained using $K-1$ portions of the generated data and tested on 1 portion of the real data to obtain the Mean Absolute Error (MAE). This process is repeated $K$ times. Similarly, by swapping the roles of the training and test sets - training on $K-1$ portions of the real data and testing on 1 portion of the generated data, we obtain another set of MAE. Ultimately, DS is calculated as the average of these two sets of MAE: 
\begin{equation}
	S = \frac{2}{K} \left( \sum_{i=1}^{K} \text{MAE}_{\text{gen-real}}(i) + \sum_{i=1}^{K} \text{MAE}_{\text{real-gen}}(i) \right).
	\label{eq9}
\end{equation}

\par In practice, we utilize GAN to generate multiple batches of generated samples, and subsequently evaluate the quality of these batches using MMD and DS. Ultimately, the best batch of generated samples is selected to be combined with real samples for training the soft sensing model. The overall pseudo code of the algorithm is presented in Algorithm \ref{RGAN-DDE}.
\begin{algorithm}[h]
	\caption{The proposed RGAN-DDE algorithm}
	\label{RGAN-DDE}
	\begin{algorithmic}[1]
		\STATE \textbf{Input:} A pool of $N$ unlabeled samples, $\{x_i\}_{i=1}^N$; The number of discriminator iterations per generator $n_{d}$; Training epoch $E$.
		\STATE \textbf{Output:} The soft sensor model $f(x)$.
		
		\STATE \textbf{Dual data evaluation strategy (training data):}
		\STATE Implement Algorithm \ref{active learning} to select high-quality training dataset $D=\{X,Y\}$ for RGAN-DDE.
		\STATE \textbf{Multi-task learning-based GAN:}
		\WHILE{$e<E$}
		\FOR{$i = 1$ to $N_{\text{critic}}$}
		\STATE Generate fake data $\{X',Y'\}$ using generator.
		\STATE Train discriminator and regressor with (\ref{eq7}).
		\STATE Update the parameters of discriminator and regressor.
		\ENDFOR
		\STATE Train generator with (\ref{eq6}).
		\STATE Update the parameters of generator.
		\ENDWHILE
		
		\STATE \textbf{Dual data evaluation strategy (generated data):}
		\STATE Generate $k$ augmented datasets $D_a=\{D_1,D_2,...,D_k\}$.
		\STATE Compute MMD and DS, and select the optimal augmented dataset $D'$. 
		\STATE Train the soft sensing model using dataset $\{D,D'\}$. 
	\end{algorithmic}
\end{algorithm}
\section{Case study}
\par In this paper, four soft sensing-related datasets demonstrate the feasibility of the proposed RGAN-DDE model. The accuracy of data augmentation and soft sensor modeling is evaluated using metrics such as Root Mean Square Error (RMSE) and Mean Absolute Error (MAE). 

\begin{table*}[t]
	\centering
	\caption{Comparison of performance enhancements for regression modeling by different generative models.}
	\label{tab:result}
	\begin{tabular}{@{}llcccccccc@{}}
		\toprule
		\textbf{Models} & \textbf{Methods}& \multicolumn{2}{c}{\textbf{Case I}} & \multicolumn{2}{c}{\textbf{Case II}}& \multicolumn{2}{c}{\textbf{Case III}} & \multicolumn{2}{c}{\textbf{Case IV}}\\
		\cmidrule(lr){3-4} \cmidrule(lr){5-6} \cmidrule(lr){7-8} \cmidrule(lr){9-10}
		& & MAE& RMSE& MAE& RMSE & MAE& RMSE & MAE& RMSE  \\
		\midrule
		&---& 0.0705& 0.0765& 0.0684& 0.0810& 0.0548& 0.0691& 0.0099&0.0116\\
		& VAE 
		& 0.0697& 0.0760& 0.0808& 0.1033& 0.0552& 0.0750& 0.0092&0.0110  \\
		& Diffusion model 
		& 0.0570& 0.0746& 0.0539& 0.0690& 0.0611& 0.0828& 0.0831&0.1044  \\
		SVR 
		& WGAN-GP 
		& 0.0403& 0.0480& 0.0428& 0.0556& 0.0352& 0.0483&0.0076& 0.0095
		\\
		& MR-GAN
		&0.0458&0.0567&0.0588&0.0733&0.0414&0.0549&0.0177&0.0193
		\\
		& RA-GAN
		&0.0355&0.0499&0.0478&0.0604&0.0416&0.0576&0.0056&0.0078
		\\
		& RGAN-DDE
		& \textbf{0.0221}& \textbf{0.0282}& \textbf{0.0332}& \textbf{0.0416}&\textbf{0.0247}& \textbf{0.0311} & \textbf{0.0047}&\textbf{0.0060} 
		\\
		\midrule
		& --- & 0.0391& 0.0664& 0.0281& 0.0344&0.0163& 0.0221 & 0.0080& 0.0113
		\\
		& VAE 
		&0.0215 & 0.0323&0.0300 & 0.0395& \textbf{0.0145}& \textbf{0.0197}& 0.0118& 0.0146 \\
		& Diffusion model
		& 0.0828& 0.1276& 0.0921& 0.1321& 0.0963& 0.1256& 0.1377&0.1759  \\
		DNN 
		& WGAN-GP 
		& 0.0343& 0.0509 & 0.0280 & 0.0346&0.0244& 0.0367 & 0.0110& 0.0150
		\\
		& MR-GAN
		&0.0239&0.0355&0.0259&0.0344&0.0147&0.0188&0.0107&0.0146
		\\
		& RA-GAN
		&\textbf{0.0206}&0.0323&0.0237&0.0301&0.0154&0.0199&0.0107&0.0141
		\\
		& RGAN-DDE
		& 0.0217& \textbf{0.0262}& \textbf{0.0220}& \textbf{0.0280}&0.0214& 0.0275& \textbf{0.0060} & \textbf{0.0078} 
		\\
		\bottomrule
	\end{tabular}
\end{table*}

\begin{table*}[t]
	\centering
	\caption{Ablation Experiment.}
	\label{tab:ablation}
	\begin{tabular}{@{}llcccccccc@{}}
		\toprule
		\textbf{Models} & \textbf{Methods}& \multicolumn{2}{c}{\textbf{Case I}} & \multicolumn{2}{c}{\textbf{Case II}}& \multicolumn{2}{c}{\textbf{Case III}} & \multicolumn{2}{c}{\textbf{Case IV}}\\
		\cmidrule(lr){3-4} \cmidrule(lr){5-6} \cmidrule(lr){7-8} \cmidrule(lr){9-10}
		& & MAE& RMSE& MAE& RMSE & MAE& RMSE & MAE& RMSE  \\
		\midrule
		&w/o shallow sharing &0.0603&0.0658&0.0377&0.0467&0.0252&0.0318&0.0061&0.0083\\
		&w/o dual data evaluation &0.0537&0.0581&0.0476&0.0594&0.0363&0.0489&0.0069&0.0089\\
		SVR &w/o DDE(train) &0.0412&0.0481&0.0415&0.0544&0.0344&0.0470&0.0056&0.0078  \\
		&w/o DDE(generated) &0.0575&0.0636&0.0338&0.0419&0.0426&0.0497&0.0072&0.0084 \\
		& RGAN-DDE
		& \textbf{0.0221}& \textbf{0.0282}& \textbf{0.0332}& \textbf{0.0416}&\textbf{0.0247}& \textbf{0.0311} & \textbf{0.0047}&\textbf{0.0060} 
		\\
		\midrule
		&w/o shallow sharing &0.0339&0.0401&0.0295&0.0363&0.0232&0.0284&0.0099&0.0133\\
		&w/o dual data evaluation &0.0244&0.0369&0.0245&0.0325&0.0174&0.0213&0.0110&0.0148\\
		DNN &w/o DDE(train) &\textbf{0.0209}&0.0300&0.0226&0.0292&\textbf{0.0156}&\textbf{0.0203}&0.0080&0.0111  \\
		&w/o DDE(generated) &0.0233&0.0292&0.0279&0.0356&0.0254&0.0316&0.0064&0.0086 \\
		& RGAN-DDE
		& 0.0217& \textbf{0.0262}& \textbf{0.0220}& \textbf{0.0280}&0.0214& 0.0275& \textbf{0.0060} & \textbf{0.0078} 
		\\
		\bottomrule
	\end{tabular}
\end{table*}

\subsection{Dataset}
\par This study employs four datasets, including two publicly available, real-world field datasets (Case I and Case III). For Case II, due to the challenges associated with collecting data from actual wastewater treatment processes, we generate the required data using the simulation platform developed by the International Water Association—an approach widely adopted by researchers in this field. Similarly, for Case IV, we generate the necessary data using a simulation platform for a carbon dioxide absorption tower.
\begin{enumerate}
	\item \textbf{Case I} - Soft Sensing of Surface Water: This case originates from the University of California, Irvine's Machine Learning Repository and comprises 25,380 samples collected from 36 sites across Georgia, USA \cite{zhao2019spatial}. Each sample includes 11 common metrics as input features, such as dissolved oxygen volume, temperature, and specific conductivity. The label represents the water quality condition for the following day. To adapt to the requirements of this study, we don't utilize all samples in the dataset; instead, we randomly select 50 samples for the training set and 200 samples for the testing set.
	\item \textbf{Case II} - Soft Sensing of Wastewater Treatment Plants: The detection of total nitrogen concentration at the effluent point of a wastewater treatment plant is crucial. However, measuring total nitrogen concentration is challenging and typically requires the use of a UV-visible spectrophotometer. To reduce costs, soft sensing is an appropriate alternative. This study employ a comprehensive wastewater treatment simulation platform developed by the International Water Association data collection. The input variables consist of 15 process variables related to wastewater treatment, while the output is the total nitrogen concentration of the effluent. The dataset includes 50 samples used as training data, 500 samples as generated data, and 200 samples as test data.
	\item \textbf{Case III} - Soft Sensing of Industrial Gas Turbines: Accurately monitoring carbon dioxide emissions from industrial gas turbines during power generation is essential for achieving energy savings, emission reductions, and environmental protection. However, real-time monitoring of carbon dioxide is very difficult, making soft sensing an appropriate alternative solution. This paper utilizes actual publicly available datasets from gas turbines in Turkey. The inputs include 9 process variables related to the engine, with the output being the carbon dioxide concentration. A total of 50 samples are used as training data, 500 samples as generated data, and 200 samples as test data.
	\item \textbf{Case IV} - Soft Sensing of Carbon Dioxide Absorption Tower: In the ammonia synthesis process, the carbon dioxide absorption tower is a critical piece of equipment \cite{9459584}. Carbon dioxide content can adversely affect the quality of the final product. However, measuring carbon dioxide concentration is challenging and typically requires the use of a mass spectrometer. To reduce measurement costs, soft sensor serves as a suitable alternative. The inputs consist of 11 variables, including pressure, flow rate, and temperature, while the output variable is the carbon dioxide concentration. A total of 50 samples are used as training data, 500 samples as generated data, and 200 samples as test data. 
	
\end{enumerate}

\begin{figure*}
	\centering
	\includegraphics[width=1\linewidth]{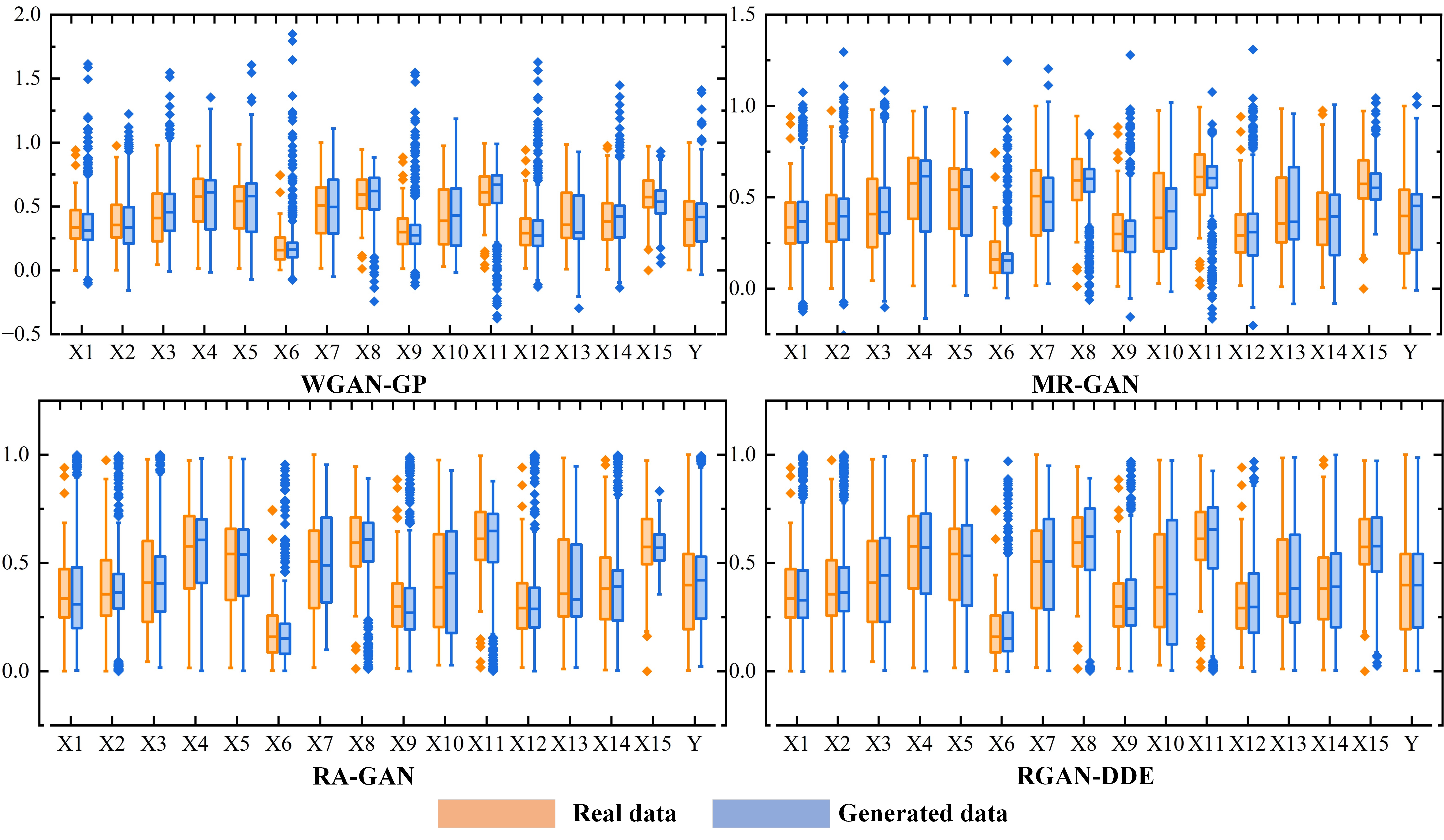}
	\caption{Visualization analysis. Boxplot comparison of real and generated feature distributions for different GAN models.}
	\label{fig:boxplot}
\end{figure*}

\begin{figure*}
	\centering
	\includegraphics[width=1\linewidth]{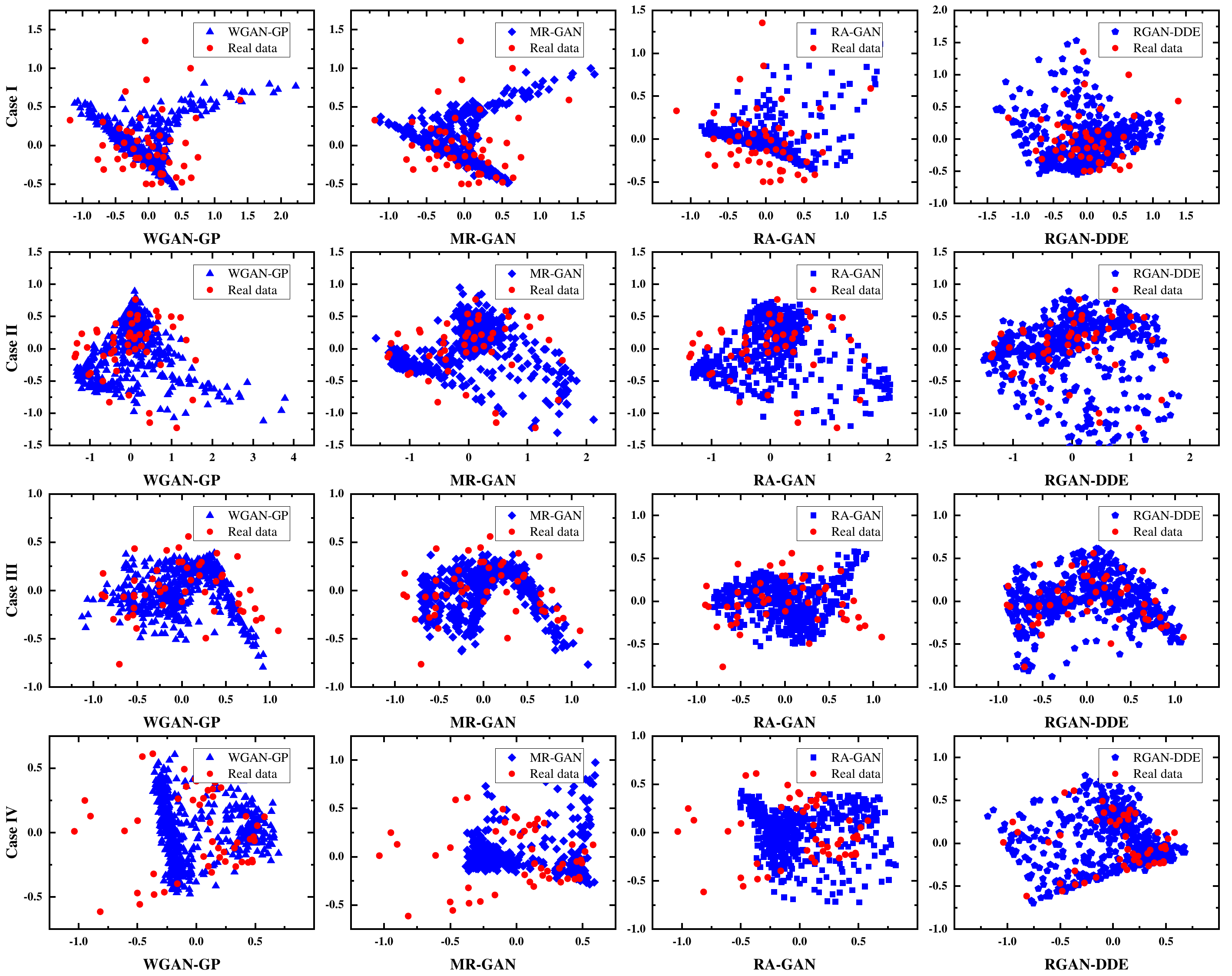}
	\caption{Visualization analysis. Scatter plot comparison of real and generated samples for four cases under different GAN-based augmentation methods.}
	\label{fig:scatter}
\end{figure*}

\subsection{Experiment Settings}
\par To evaluate the performance of the proposed RGAN-DDE, SOTA algorithms are selected for comparison, including: 1) VAE \cite{xie2019supervised}; 2) Diffusion model \cite{liu2024flame}; 3) WGAN-GP \cite{gulrajani2017improved}; 4) MR-GAN \cite{li2023data}; 5) RA-GAN \cite{jiang2022ragan}; 6) RGAN-DDE. WGAN-GP is the benchmark method, while MR-GAN and RA-GAN are improved methods that take into account the mapping relationship between dependent and independent variables. RGAN-DDE is the methods proposed in this paper. The architectures for both the generator and discriminator in all models are set to 32-32. For MR-GAN, RA-GAN and RGAN-DDE, the regressor or regression attention architecture is configured as 32-8. The activation function used throughout is LeakyReLU. The optimizer algorithm is Adam. The penalty weight \(\lambda\), the number of critics, the dimension of the noise vector, the learning rate, batch size, and iteration times of training are set to 0.5, 5, 8, 0.0001, 32, and 10000, respectively.

\begin{figure}
	\centering
	\includegraphics[width=\linewidth]{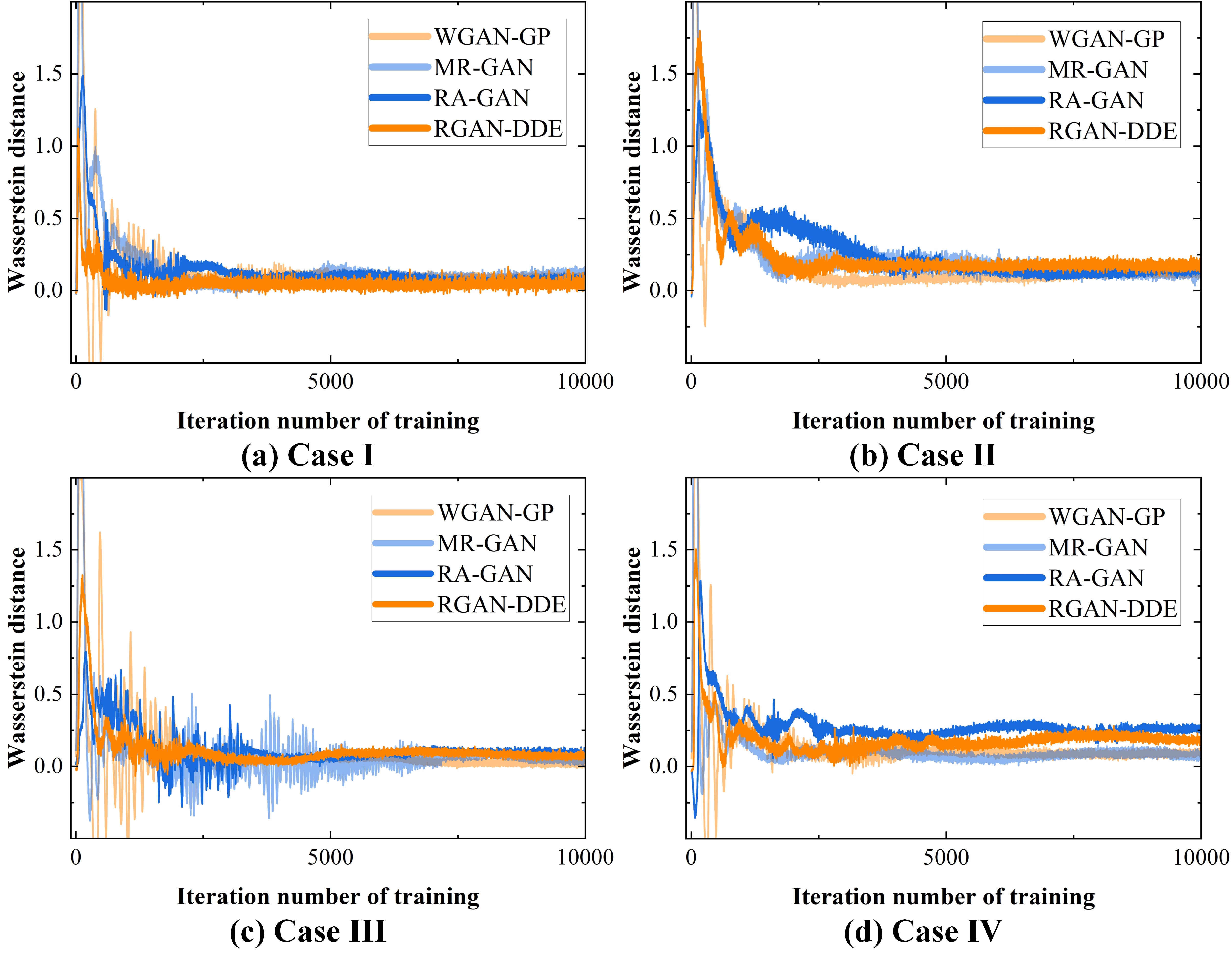}
	\caption{Wasserstein distance of different generative models}
	\label{fig:convergence}
\end{figure}
\subsection{Main Results}

\par We apply data augmentation techniques to soft sensor modeling to investigate their effectiveness in enhancing modeling accuracy. Support Vector Regression (SVR) and Deep Neural Networks (DNN) are selected as representative regression models to evaluate the impact of data augmentation on both linear (SVR) and nonlinear (DNN) modeling frameworks. The key conclusions derived from the Table \ref{tab:result} are as follows:
\par First, the proposed RGAN‑DDE achieves the best performance in 13 out of the 16 evaluation metrics.  For SVR, RGAN‑DDE consistently yields the lowest MAE and RMSE across all four cases, reducing the MAE by approximately 50\% relative to the non‑augmented baseline. For DNN, RGAN‑DDE also provides substantial gains in three out of four cases, with only a slight degradation observed in Case III where the baseline model already exhibits very low error. These results indicate that, by incorporating shallow‑layer feature sharing between the discriminator and regressor and adopting a dual data quality evaluation mechanism, RGAN‑DDE can generate augmented samples that are both distributionally consistent with the historical data and sufficiently diverse. Consequently, it markedly enhances the accuracy and generalization ability of downstream soft sensor models, thereby alleviating the small‑sample issue that is pervasive in industrial applications.
\par In contrast, VAE, diffusion models and WGAN‑GP do not explicitly consider the regression relationship between process variables and target variables during sample generation, which leads to inferior or even detrimental effects in several scenarios. For SVR, the diffusion model and VAE offers only marginal improvements or no gain. For DNN, diffusion models again perform poorly in multiple cases, which can be attributed to their high sensitivity to network architecture and hyper‑parameters. Nevertheless, a more nuanced pattern can be observed: WGAN‑GP provides a noticeable improvement for SVR but only limited benefit for DNN, whereas VAE substantially enhances DNN performance while offering limited gains for SVR. A plausible explanation is that WGAN‑GP is more prone to generating samples that follow relatively simple, linear relationships, which align better with the hypothesis space of SVR, whereas VAE is more effective at capturing highly nonlinear variability, which can be better exploited by DNN.
\par MR-GAN and RA-GAN achieve the second-best overall performance, attributable to their explicit consideration of the relationship between independent and dependent variables. MR-GAN excels at enhancing DNN performance but exhibits inferior performance with SVR. This is likely because its iterative training process repeatedly updates the regressor, potentially causing generated samples to overfit nonlinear patterns. RA-GAN, which utilizes a frozen regressor, exhibits performance slightly inferior to RGAN-DDE.

\begin{table}[t]
	\centering
	\caption{Comparison of running times of different generation models.}
	\label{tab:efficiency}
	
	\begin{tabular}{@{}lcccc@{}}
		\toprule
		\textbf{Methods}&{\textbf{Case I}} & {\textbf{Case II}}&{\textbf{Case III}} & {\textbf{Case IV}}\\
		
		\midrule
		VAE&47.1&43.1&42.8&45.1\\
		Diffusion model&45.9&41.3&41.3&40.9\\
		WGAN-GP&359.6&334.7&351.9&367.7\\
		MR-GAN&21176.3&21567.2&30322.3&21693.4\\
		RA-GAN&733.1&694.9&672.1&704.5\\
		RGAN-DDE&673.7&688.8&708.9&700.3\\
		\bottomrule
	\end{tabular}
\end{table}

\subsection{Ablation Experiment}

\par In this section, we investigate the role of each component in RGAN-DDE. The first setting removes the shallow-sharing mechanism of RGAN-DDE, eliminating parameter sharing between the discriminator and the regressor. The regressor is trained only initially and then its parameters are frozen. The second setting removes the dual data evaluation strategy, meaning no selection of training data or generated samples is performed. To further examine the effect of the dual data evaluation strategy, we separately consider selecting only training samples or only generated samples, to identify which part more effectively guides the GAN's generation direction. The ablation study results are shown in the Table \ref{tab:ablation}. The shallow sharing mechanism and generated sample evaluation in RGAN-DDE consistently have a positive effect across all cases, contributing to improved accuracy in soft sensing. However, it is worth noting that selecting training samples occasionally has a negative impact. This may be related to the active learning algorithm chosen in this study.

\begin{figure*}
	\centering
	\includegraphics[width=1\linewidth]{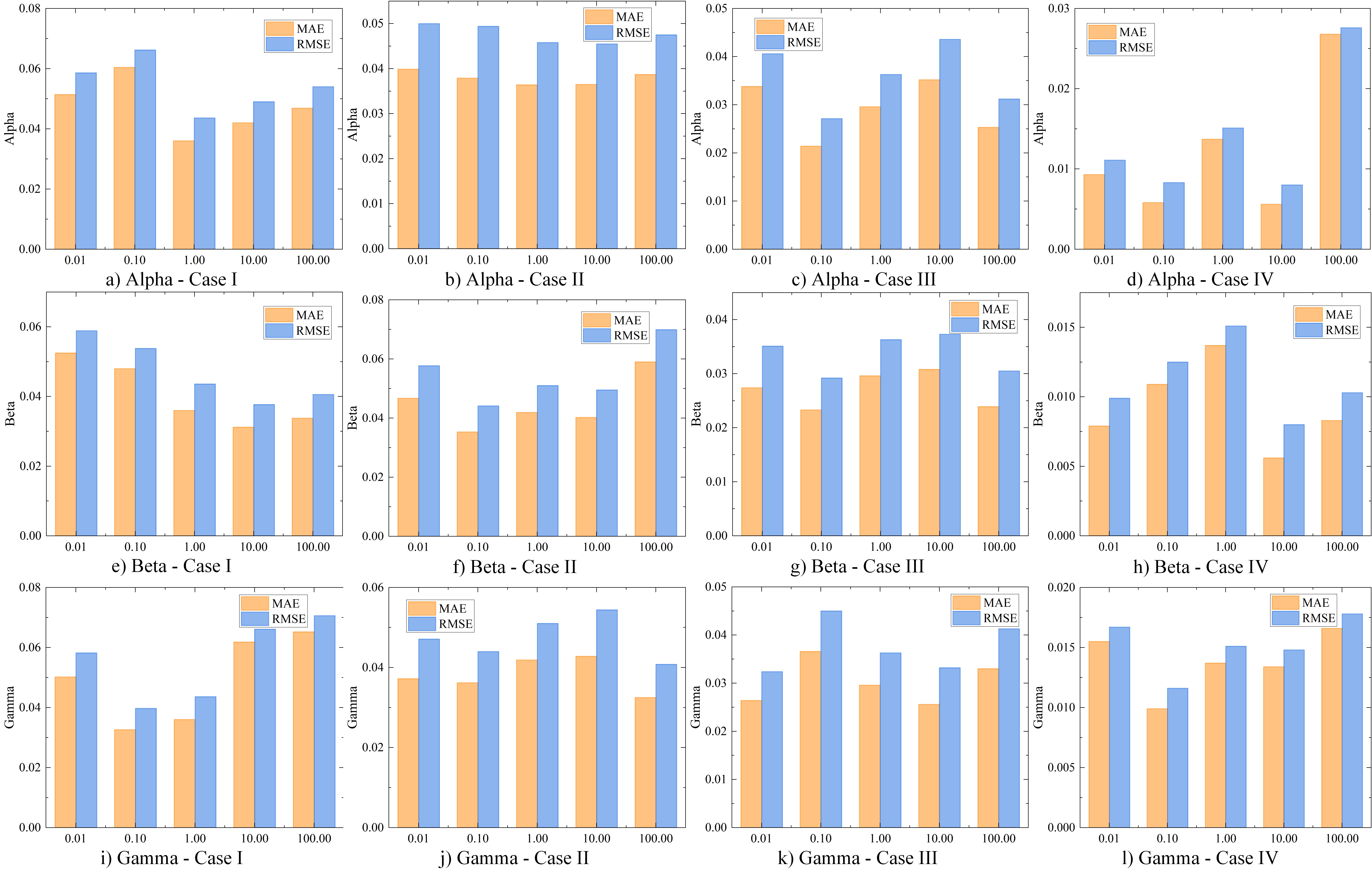}
	\caption{Hyper-parameter sensituvity analysis.}
	\label{fig:hyper}
\end{figure*}

\subsection{Model Analysis}

\par \textbf{Visualization analysis}. To intuitively assess the similarity between the distributions of the generated data and the real data, we present in Fig. \ref{fig:boxplot} the boxplot comparisons of four generative models (WGAN‑GP, MR‑GAN, RA‑GAN, and RGAN‑DDE) across all feature dimensions for Case II. Overall, all models capture the basic trends of the real data to some extent. However, they exhibit notable differences in the accuracy and stability of distribution fitting. RGAN‑DDE shows the closest match to the real distribution, followed by RA‑GAN and MR‑GAN, while WGAN‑GP displays more pronounced deviations.
\par WGAN‑GP shows noticeable shifts between the generated and real data in most dimensions and produces many outliers. This indicates that the data generated by WGAN‑GP contain considerable noise, and the model struggles to constrain extreme values, leading to a pronounced long‑tail effect in the generated distribution. MR‑GAN and RA‑GAN show clear improvements over WGAN‑GP. The number of outliers decreases, and the overlap between boxplots increases, suggesting that incorporating regression information enhances the models’ ability to suppress noisy samples. However, both models still show median shifts in a few dimensions (especially Y). MR‑GAN exhibits slight range shrinkage in X2, X3, X6, X7, X8, X11, and X15, while RA‑GAN shows similar shrinkage in X2, X3, X8, and X15. RGAN‑DDE achieves almost complete overlap between the blue and orange boxplots across nearly all dimensions, with whisker lengths and outlier patterns closely matching the real data. This includes the Y dimension, where the generated data not only reproduce the overall range but also accurately capture extreme tail values, which is particularly important for subsequent Y‑based regression tasks. A minor shrinkage remains in the X2 dimension.
\par In addition, we project the high-dimensional samples onto a two-dimensional plane to more intuitively compare the performance of RGAN‑DDE with other baseline methods. As shown in Fig. \ref{fig:scatter}, each row corresponds to a different dataset, each column corresponds to a different algorithm, blue denotes the generated data, and orange denotes the real training data.
\par  Next, we focus on Case I and Case IV, which are representative in terms of generation quality. For Case I, the samples generated by WGAN‑GP, MR‑GAN, and RA‑GAN are mainly concentrated in a narrow band, and they fail to adequately capture the region in the lower-left corner where real samples are densely distributed, leaving part of the data manifold uncovered. In contrast, the point cloud generated by RGAN‑DDE is more uniformly distributed across the feature space and closely matches both the global shape and local structure of the real point cloud. It not only covers the hard‑to‑fit lower-left region but also significantly improves sample diversity. For Case IV, WGAN‑GP and MR‑GAN still produce many drifting points outside the high‑density area, with noticeable mode collapse in some regions. RA‑GAN generates samples more concentrated in the true data region, but slight expansion and leakage can still be observed near the boundaries. In comparison, RGAN‑DDE forms a point cloud that almost completely overlaps with the real data in the high‑density core area, highlighting the advantages brought by incorporating regression information and data‑quality discrimination.

\begin{table*}[t]
	\centering
	\caption{Results of RGAN-DDE with different data.}
	\label{tab:amount}
	\begin{tabular}{@{}lccccccccc@{}}
		\toprule
		& & \multicolumn{2}{c}{\textbf{Case I}} & \multicolumn{2}{c}{\textbf{Case II}}& \multicolumn{2}{c}{\textbf{Case III}} & \multicolumn{2}{c}{\textbf{Case IV}}\\
		\cmidrule(lr){3-4} \cmidrule(lr){5-6} \cmidrule(lr){7-8} \cmidrule(lr){9-10}
		\textbf{Models} & \textbf{Amount of generated data} & MAE& RMSE& MAE& RMSE & MAE& RMSE & MAE& RMSE  \\
		\midrule
		&100 &0.0552&0.0608&0.0425&0.0515&0.0346&0.0430&0.0082&0.0101\\
		&200 &0.0481&0.0531&0.0395&0.0491&0.0328&0.0409&0.0068&0.0089\\
		SVR &300 &0.0433&0.0514&0.0390&0.0480&0.0314&0.0394&0.0068&0.0089  \\
		&400 &0.0411&0.0463&0.0379&0.0468&0.0341&0.0430&0.0068&0.0089 \\
		& 500&0.0378&0.0428&0.0371&0.0456&0.0348&0.0428&\textbf{0.0065}&\textbf{0.0087} \\
		&1000&\textbf{0.0332}&\textbf{0.0394}&\textbf{0.0366}&\textbf{0.0454}&\textbf{0.0312}&\textbf{0.0389}&0.0069&0.0090\\
		\midrule
		&100 &0.0259&0.0337&0.0285&0.0386&0.0188&0.0241&0.0066&0.0094\\
		&200 &\textbf{0.0229}&0.0309&0.0265&0.0346&0.0170&0.0211&0.0071&0.0101\\
		DNN &300 &0.0244&0.0351&\textbf{0.0256}&\textbf{0.0317}&\textbf{0.0145}&\textbf{0.0177}&0.0071&0.0088  \\
		&400 &0.0235&\textbf{0.0299}&0.0274&0.0352&0.0155&0.0195&0.0080&0.0103 \\
		& 500&0.0248&0.0346&0.0284&0.0355&0.0155&0.0195&\textbf{0.0060}&\textbf{0.0079} \\
		&1000&0.0271&0.0332&0.0275&0.0356&0.0166&0.0204&0.0073&0.0095
		\\
		\bottomrule
	\end{tabular}
\end{table*}

\par \textbf{Efficiency analysis}. We evaluate the computational efficiency and convergence of various generative algorithms. Computational efficiency is assessed by training time, with results shown in the Table \ref{tab:efficiency}. Convergence is evaluated using the Wasserstein distance curve, with results presented in the Figure \ref{fig:convergence}. As shown in the Table \ref{tab:efficiency}, VAE and diffusion models have the shortest running times. This is due to our adoption of a relatively simple design and the fact that they do not model the regression relationship between dependent and independent variables, which further reduces computational time. Clearly, the proposed method, RGAN-DDE, achieves shorter running time among the three generative models that consider regression relationships. This is attributed to our shallow shared mechanism, which allows RGAN-DDE to balance performance and efficiency. In contrast, MR-GAN updates its regressor in every iteration, resulting in significantly longer running time compared to other methods. As shown in the Figure \ref{fig:convergence}, RGAN-DDE achieves the best convergence curves across all cases — especially in Case 3, where other methods exhibit significant oscillations, while RGAN-DDE converges rapidly and stably.

\par \textbf{Hyperparameter sensitivity analysis}. We set $\alpha$, $\beta$, and $\gamma$ to their default value of 1, then varied one parameter at a time over the range [0.01, 0.1, 1, 10, 100]. As shown in the Figure \ref{fig:hyper}, there is no universal parameter setting that achieves optimal performance for RGAN-DDE across all datasets. However, it is clear that both very large and very small parameter values lead to performance degradation. This indicates that each regularization term in the generator and discriminator’s objective functions contributes positively to overall performance. Ultimately, the optimal values for $\alpha$, $\beta$, and $\gamma$ were set to [1, 10, 0.1] in Case I, [1, 0.1, 100] in Case II, [0.1, 0.1, 10] in Case III, and [10, 10, 0.1] in Case IV. Default values of $\alpha$, $\beta$, and $\gamma$ may be selected as 0.1 or 1, under which the system performance remains satisfactory.

\par \textbf{Others}. Considering the significant impact of the quantity of generated data on the accuracy of regression modeling, we conduct comparative experiments on the amount of data generated by RGAN-DDE. As shown in the Table \ref{tab:amount}, in most cases, the accuracy of the regression model first increases and then decreases as the number of generated samples grows — indicating an optimal amount of synthetic data. In Case 4, however, accuracy increases initially and then stabilizes. We conclude that insufficient data limits model performance due to lack of information, while excessive data may introduce noisy or anomalous samples that degrade accuracy. This study focuses only on selecting more diverse generated samples, and does not address how to determine the optimal number of samples — a direction we plan to explore in future work.

\section{Conclusion}
\par This paper proposes a data augmentation method named RGAN-DDE, aimed at addressing the few-shot sample problem in industrial soft sensing. Firstly, the mapping relationship between labels and features is embedded into the generator and discriminator to further enhance the quality of generated samples. Subsequently, considering the similarity between the regressor and the discriminator, shallow sharing is implemented between them, ensuring generation quality while accelerating convergence speed. Finally, recognizing the importance of both training samples and generated samples, a dual data evaluation strategy is proposed to guide the model in generating more diverse samples, thereby increasing the generalization ability of subsequent modeling. The proposed method is validated using typical industrial soft sensing cases and compared with other advanced methods. Experimental results demonstrate the superiority of RGAN-DDE, significantly enhancing the accuracy and generalization ability of subsequent modeling. However, this study does not consider the issue of the optimal amount of generated data and the best GAN structure, which are one of the important directions for future research.

\vspace{1em}
\noindent\textbf{Acknowledge}
\par This work is supported in part by the National Natural Science Foundation of China under Grant 62394341, Grant 62027811 and Grant 618909302. The authors acknowledge the computational resources provided by the High-Performance Computing Centre of Central South University, China. 

% Use the style of numbering in square brackets.
% If nothing is used, default style will be taken.
%\begin{enumerate}[a)]
%\item 
%\item 
%\item 
%\end{enumerate}  

%\begin{itemize}
%\item 
%\item 
%\item 
%\end{itemize}  

%\begin{description}
%\item[]
%\item[] 
%\item[] 
%\end{description}  

% Figure
% \begin{figure}[<options>]
% 	\centering
% 		\includegraphics[<options>]{}
% 	  \caption{}\label{fig1}
% \end{figure}

% \begin{table}[<options>]
% \begin{tabular*}{\tblwidth}{@{}LL@{}}
% \toprule
% \midrule
%  & \\
%  & \\
%  & \\
%  & \\
% \bottomrule
% \end{tabular*}
% \end{table}

% Uncomment and use as the case may be
%\begin{theorem} 
%\end{theorem}

% Uncomment and use as the case may be
%\begin{lemma} 
%\end{lemma}



\bibliographystyle{elsarticle-num}

\bibliography{ref}

\end{document}